\documentclass[journal]{IEEEtran}
\usepackage{cite}

\ifCLASSINFOpdf
 \usepackage[pdftex]{graphicx}
\else
\fi
\usepackage[cmex10]{amsmath}
\usepackage{atbegshi}
\AtBeginDocument{\AtBeginShipoutNext{\AtBeginShipoutDiscard}}

\usepackage{nicefrac}
\usepackage{amsfonts}
\usepackage{amsthm}

\newtheorem{mylemma}{Lemma}

\begin{document}
%
\title{Backpropagation generalized for output derivatives}
%
%
%

\author{V.I. Avrutskiy}
\thanks{V.I. Avrutskiy is with the Department of Aeromechanics and Flight Engineering of Moscow Institute of Physics and Technology, Institutsky lane 9, Dolgoprudny, Moscow region, 141700, e-mail: avrutsky@phystech.edu}

\maketitle

\begin{abstract}
Backpropagation algorithm is the cornerstone for neural network analysis. Paper extends it for training any derivatives of neural network's output with respect to its input. By the dint of it feedforward networks can be used to solve or verify solutions of partial or simple, linear or nonlinear differential equations. This method vastly differs from traditional ones like finite differences on a mesh. It contains no approximations, but rather an exact form of differential operators. Algorithm is built to train a feed forward network with any number of hidden layers and any kind of sufficiently smooth activation functions. It's presented in a form of matrix-vector products so highly parallel implementation is readily possible. First part derives the method for 2D case with first and second order derivatives, second part extends it to N-dimensional case with any derivatives. All necessary expressions for using this method to solve most applied PDE can be found in Appendix D.
\end{abstract}


%

\section{Introduction}
%
%
%
%


Backpropagation procedure\cite{1,2} proved itself as a very powerful tool for weights training. Most of the problems successfully solved by neural networks could not be approached without it. Since then, feedforward networks were used a lot as universal function approximators\cite{3,4,5,6}. And as a consequence of fitting values they also fit function's derivatives\cite{7}. This feature allowed to extend the use of neural networks and successfully solve ordinary and partial differential equations\cite{8,9,10,11,12,13} using various optimization techniques.

Applying neural networks to solve PDE is a totally different practice compared to classical methods. It nearest analog is the undetermined coefficients method. It consists of substitution of a certain parametric function into differential equation, after which parameters are found as a solution of an ordinary equation. In this case, all weights and thresholds of neural network are undetermined coefficients which defining equation can be solved via gradient methods. Unlike finite differences which work with approximated differential operators (numerical viscosity being one of the possible consequences) training procedure for neural network relies on the exact value of field derivatives. Training algorithms like gradient descent or RProp\cite{14} cannot diverge or oscillate. When representing a sufficiently smooth function, neural network uses relatively small amount of memory compared to that required for storing values defined on a mesh.

So far, only limited number of feedforward network architectures was used for solving PDE. Most of them being radial basis networks with one hidden layer\cite{15,16,17,18,19}. Some impressive results for training based on matrix multiplications were obtained for a particular case with first order derivative and three layer network\cite{20}.

Providing more freedom to neural network's architecture was a key motivator for this paper. Another important generalization is a single technique for any derivatives used in PDE. Article will be focused only on obtaining derivatives of an error function with respect to weights. When calculated, they can be used in a number of developed methods such as RProp or gradient descent.
\subsection{Notation}
Connections between neighbor layers are gathered into matrices so that $W^{\alpha_{2}\;\alpha_{1}}_{(L_{2})(L_{1})}$, connects $\alpha_{1}$ neuron of layer $L_{1}$ to $\alpha_{2}$ neuron of $L_{2}$. Signal is propagated from $L_{1}$ to $L_{2}$. Input values are considered as layer 0. Activation function for every neuron in each layer is denoted as $\sigma$. Generalization for various functions is straightforward. Signals for input neurons are denoted by $a$ and $b$.
\begin{figure*}
\begin{center}
 \includegraphics{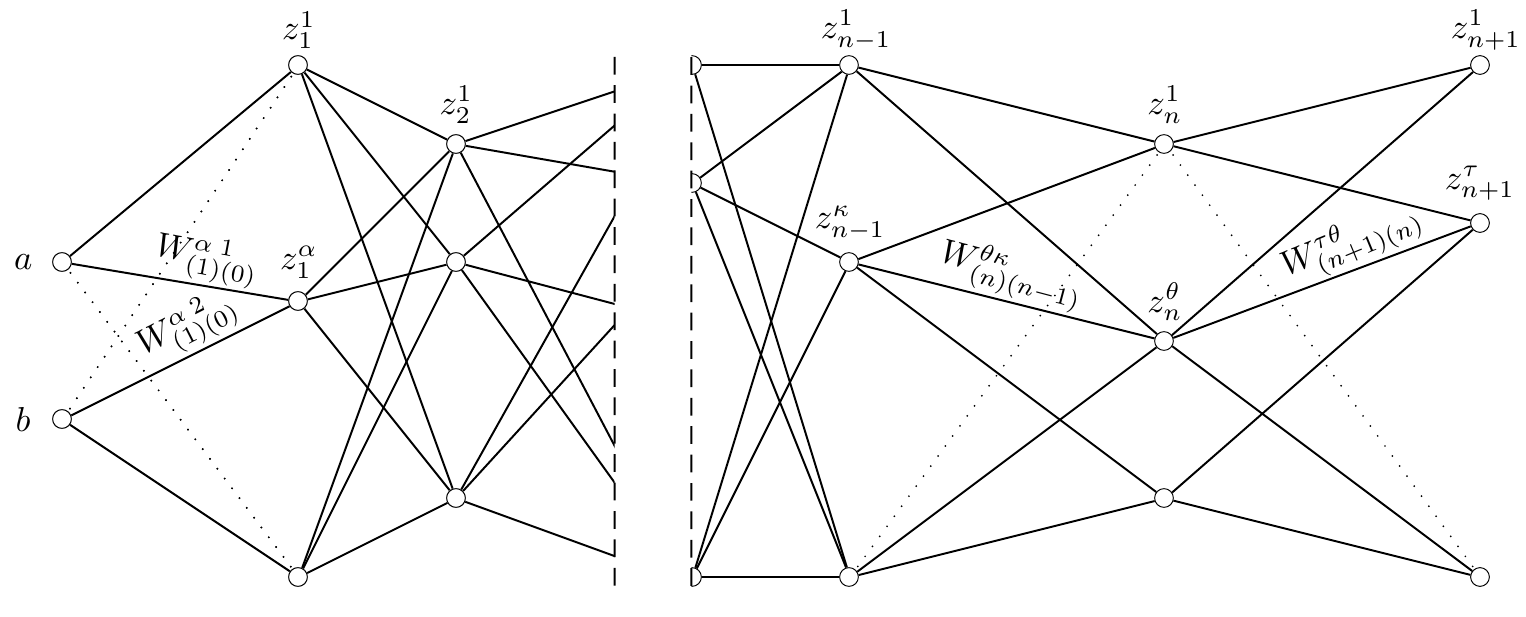}
 \end{center}
 \end{figure*}

\section{Two dimensional case, second order}
\subsection{Forward propagation}
Obtaining derivatives of output with respect to input requires an extension of feedforward procedure. Those derivatives will propagate from one layer to another in the similar way as values or neural network are calculated. Initial conditions for forward pass are calculated directly from the first layer definition:
\begin{equation}z_{1}^{\alpha}=W_{(1)(0)}^{\alpha\,1}a+W_{(1)(0)}^{\alpha\,2}b \label{initVals} \end{equation}
\begin{equation}\frac{\partial z_{1}^{\alpha}}{\partial a}=W_{(1)(0)}^{\alpha\,1}\quad\frac{\partial z_{1}^{\alpha}}{\partial b}=W_{(1)(0)}^{\alpha\,2} \label{initFirst}\end{equation}
\begin{equation}\frac{\partial^{2}z_{1}^{\alpha}}{\partial a^{2}}=0\quad\frac{\partial^{2}z_{1}^{\alpha}}{\partial b^{2}}=0\quad\frac{\partial^{2}z_{1}^{\alpha}}{\partial a\partial b}=0\label{initSecond} \end{equation}
Recurrent relation for calculating next layer values from previous:
\begin{equation}
z_{n}^{\theta}=\sum_{\kappa}W_{(n)(n-1)}^{\theta\kappa}\sigma\left(z_{n-1}^{\kappa}+t_{n-1}^{\kappa}\right)
\label{valsProp}
\end{equation}
We will omit threshold $t$ in further expressions.
Using the chain rule, derivative or hidden layer values with respect to $a$ can be written as a matrix-vector product:
\[\frac{\partial z_{n}^{\theta}}{\partial a}=\sum_{\kappa}\frac{\partial z_{n}^{\theta}}{\partial z_{n-1}^{\kappa}}\frac{\partial z_{n-1}^{\kappa}}{\partial a}=\left(\frac{\partial z_{n}^{\theta}}{\partial z_{n-1}^{\kappa}}\right)\cdot\left(\frac{\partial z_{n-1}^{\kappa}}{\partial a}\right)\]
where the matrix according to (\ref{valsProp}) is:
\[\frac{\partial z_{n}^{\theta}}{\partial z_{n-1}^{\kappa}}=W_{(n)(n-1)}^{\theta\kappa}\sigma'\left(z_{n-1}^{\kappa}\right)\]
Substituting it gives us a formula for first derivative forward pass:
\begin{equation}
\frac{\partial z_{n}^{\theta}}{\partial a}=\sum_{\kappa}W_{(n)(n-1)}^{\theta\kappa}\sigma'\left(z_{n-1}^{\kappa}\right)\frac{\partial z_{n-1}^{\kappa}}{\partial a}
\label{firstProp}
\end{equation}
Applying the chain rule once again to obtain second derivative:
\begin{gather*}
\frac{\partial^{2}z_{n}^{\theta}}{\partial a^{2}}=\frac{\partial}{\partial a}\left(\frac{\partial z_{n}^{\theta}}{\partial z_{n-1}^{\kappa}}\right)\cdot\left(\frac{\partial z_{n-1}^{\kappa}}{\partial a}\right)+\notag\\
\left(\frac{\partial z_{n}^{\theta}}{\partial z_{n-1}^{\kappa}}\right)\cdot\frac{\partial}{\partial a}\left(\frac{\partial z_{n-1}^{\kappa}}{\partial a}\right)
\end{gather*}
\[\frac{\partial}{\partial a}\left(\frac{\partial z_{n}^{\theta}}{\partial z_{n-1}^{\kappa}}\right)=\sum_{\kappa'}\frac{\partial^{2}z_{n}^{\theta}}{\partial z_{n-1}^{\kappa'}\partial z_{n-1}^{\kappa}}\frac{\partial z_{n-1}^{\kappa'}}{\partial a}\]
And noting from (\ref{valsProp}) that variables from $n$th layer have zero mixed derivatives with respect to $n-1$st:
\begin{equation*}
\frac{\partial^{2}z_{n}^{\theta}}{\partial z_{n-1}^{\kappa'}\partial z_{n-1}^{\kappa}}=\left(\frac{\partial^{2}z_{n}^{\theta}}{(\partial z_{n-1}^{\kappa})^{2}}\right)\delta_{\kappa'\kappa}
\end{equation*}
second derivative on next layer can be found from simple expression:
\begin{gather*}
\frac{\partial^{2}z_{n}^{\theta}}{\partial a^{2}}=\left(\frac{\partial^{2}z_{n}^{\theta}}{(\partial z_{n-1}^{\kappa})^{2}}\right)\cdot\left(\frac{\partial z_{n-1}^{\kappa}}{\partial a}\right)^{2}+\notag\\
+\left(\frac{\partial z_{n}^{\theta}}{\partial z_{n-1}^{\kappa}}\right)\cdot\left(\frac{\partial^{2}z_{n-1}^{\kappa}}{\partial a^{2}}\right)
\end{gather*}
which upon substituting two ``layer-to-layer'' matrices:
\[\frac{\partial z_{n}^{\theta}}{\partial z_{n-1}^{\kappa}}=W_{(n)(n-1)}^{\theta\kappa}\sigma'\left(z_{n-1}^{\kappa}\right)\]
\[\frac{\partial^{2}z_{n}^{\theta}}{(\partial z_{n-1}^{\kappa})^{2}}=W_{(n)(n-1)}^{\theta\kappa}\sigma''\left(z_{n-1}^{\kappa}\right)\]
turns into a final forward pass formula for second derivative:
\begin{gather}
\frac{\partial^{2}z_{n}^{\theta}}{\partial a^{2}}=\sum_{\kappa}W_{(n)(n-1)}^{\theta\kappa}\left(\sigma''\left(z_{n-1}^{\kappa}\right)\left(\frac{\partial z_{n-1}^{\kappa}}{\partial a}\right)^{2}+\right.\notag\\
\left. \vphantom{\left(\frac{\partial z_{n-1}^{\kappa}}{\partial a}\right)^{2}} + \sigma'\left(z_{n-1}^{\kappa}\right)\frac{\partial^{2}z_{n-1}^{\kappa}}{\partial a^{2}}\right)
\label{secondProp}
\end{gather}
Expression for mixed derivative is quite similar:
\begin{gather}
\frac{\partial^{2}z_{n}^{\theta}}{\partial a\partial b}=\sum_{\kappa}W_{(n)(n-1)}^{\theta\kappa}\!\left(\sigma''\left(z_{n-1}^{\kappa}\right)\!\left(\frac{\partial z_{n-1}^{\kappa}}{\partial a}\,\frac{\partial z_{n-1}^{\kappa}}{\partial b}\right)+\right.\notag\\
\left.+\sigma'\left(z_{n-1}^{\kappa}\right)\frac{\partial^{2}z_{n-1}^{\kappa}}{\partial a\partial b}\right)
\label{mixedProp}
\end{gather}
Formula (\ref{valsProp}) for forward pass is now completed with (\ref{firstProp}) and (\ref{secondProp}) for derivatives with respect to $a$ (and similar formulas with respect to $b$) along with (\ref{mixedProp}) for mixed derivative. Now we are able to calculate error function which is extended to contain not only exit layer values but also their derivatives  with respect to the input.
\subsection{Backward propagation}
In order to simplify the overall mathematical narrative we will use dots to denote partial derivatives with respect to $a$, omitting any dependency of error function from $b$ and related partials. For $n+1$st layer:
\begin{equation}\dot{z}_{n+1}^{\tau}\equiv\frac{\partial z_{n+1}^{\tau}}{\partial a}\qquad\ddot{z}_{n+1}^{\tau}\equiv\frac{\partial^{2}z_{n+1}^{\tau}}{\partial a^{2}}\label{lastDeriv}\end{equation}
\begin{equation}z_{n+1}^{\tau}=\sum_{\theta}W_{(n+1)(n)}^{\tau\theta}\sigma\left(z_{n}^{\theta}\right)\label{lastVals}\end{equation}
Considering (\ref{lastVals}) and (\ref{lastDeriv}) as the output (thus last layer is linear) we evaluate the error function, which in new notations is written as:
\[E\left(a,z_{n+1}^{\tau},\dot{z}_{n+1}^{\tau},\ddot{z}_{n+1}^{\tau}\right)\!\equiv E\left(\!a,z_{n+1}^{\tau},\frac{\partial z_{n+1}^{\tau}}{\partial a},\frac{\partial^{2}z_{n+1}^{\tau}}{\partial a^{2}}\!\right)\]
and its partial derivatives with respect to output values and their derivatives:
\begin{equation}\frac{\partial E}{\partial z_{n+1}^{\tau}},\frac{\partial E}{\partial\dot{z}_{n+1}^{\tau}},\frac{\partial E}{\partial\ddot{z}_{n+1}^{\tau}}\label{eDerivLast}\end{equation}
Which allow us to calculate derivatives related to last weights matrix $W_{(n+1)(n)}^{\tau\theta}$ using the chain rule:
\begin{gather}
\frac{\partial E}{\partial W_{(n+1)(n)}^{\tau\theta}}=\sum_{\tau'}\frac{\partial E}{\partial z_{n+1}^{\tau'}}\frac{\partial z_{n+1}^{\tau'}}{\partial W_{(n+1)(n)}^{\tau\theta}}+\notag\\
\frac{\partial E}{\partial\dot{z}_{n+1}^{\tau'}}\frac{\partial\dot{z}_{n+1}^{\tau'}}{\partial W_{(n+1)(n)}^{\tau\theta}}+\frac{\partial E}{\partial\ddot{z}_{n+1}^{\tau'}}\frac{\partial\ddot{z}_{n+1}^{\tau'}}{\partial W_{(n+1)(n)}^{\tau\theta}}\label{lastWeightsBasic}
\end{gather}
With (\ref{valsProp}),(\ref{firstProp}) and (\ref{secondProp}) expressions written for the last layer one can get all necessary relations:
\[\frac{\partial z_{n+1}^{\tau'}}{\partial W_{(n+1)(n)}^{\tau\theta}}=\sigma\left(z_{n}^{\theta}\right)\delta_{\tau'\tau}\]
\[\frac{\partial\dot{z}_{n+1}^{\tau'}}{\partial W_{(n+1)(n)}^{\tau\theta}}=\sigma'\left(z_{n}^{\theta}\right)\dot{z}_{n}^{\theta}\:\delta_{\tau'\tau}\]
\[\frac{\partial\ddot{z}_{n+1}^{\tau'}}{\partial W_{(n+1)(n)}^{\tau\theta}}=\left(\sigma''\left(z_{n}^{\theta}\right)\left(\dot{z}_{n}^{\theta}\right)^{2}+\sigma'\left(z_{n}^{\theta}\right)\ddot{z}_{n}^{\theta}\right)\delta_{\tau'\tau}\]
And substitute them into (\ref{lastWeightsBasic}) to get error derivatives with respect to weights:
\begin{gather}
\frac{\partial E}{\partial W_{(n+1)(n)}^{\tau\theta}}=\frac{\partial E}{\partial z_{n+1}^{\tau}}\sigma\left(z_{n}^{\theta}\right)+\frac{\partial E}{\partial\dot{z}_{n+1}^{\tau}}\sigma'\left(z_{n}^{\theta}\right)\dot{z}_{n}^{\theta}+\notag\\
+\frac{\partial E}{\partial \ddot{z}_{n+1}^{\tau}}\left(\sigma''\left(z_{n}^{\theta}\right)\left(\dot{z}_{n}^{\theta}\right)^{2}+\sigma'\left(z_{n}^{\theta}\right)\ddot{z}_{n}^{\theta}\right)
\label{lastWeightsFull}
\end{gather}
Now the goal is to propagate expressions (\ref{eDerivLast}) backwards to $n$th layer, considering $E$ as a function of previous layer variables and their derivatives with respect to input:
\[E\left(a,z_{n+1}^{\tau},\dot{z}_{n+1}^{\tau},\ddot{z}_{n+1}^{\tau}\right)\rightarrow E\left(a,z_{n}^{\theta},\partial\dot{z}_{n}^{\theta},\partial\ddot{z}_{n}^{\theta}\right)\]
This can be done using the chain rule. For $n$th layer values:
\begin{gather}
\frac{\partial E}{\partial z_{n}^{\theta}}=\sum_{\tau}\frac{\partial E}{\partial z_{n+1}^{\tau}}\frac{\partial z_{n+1}^{\tau}}{\partial z_{n}^{\theta}}+\frac{\partial E}{\partial\dot{z}_{n+1}^{\tau}}\frac{\partial\dot{z}_{n+1}^{\tau}}{\partial z_{n}^{\theta}}+\notag\\
+\frac{\partial E}{\partial\ddot{z}_{n+1}^{\tau}}\frac{\partial\ddot{z}_{n+1}^{\tau}}{\partial z_{n}^{\theta}}\label{bpVals}
\end{gather}
with necessary parts obtained from (\ref{valsProp}), (\ref{firstProp}) and (\ref{secondProp}):
\begin{equation}
\frac{\partial z_{n+1}^{\tau}}{\partial z_{n}^{\theta}}=W_{(n+1)(n)}^{\tau\theta}\sigma'\left(z_{n}^{\theta}\right)\label{valsOvals}
\end{equation}
\begin{equation}
\frac{\partial\dot{z}_{n+1}^{\tau}}{\partial z_{n}^{\theta}}=W_{(n+1)(n)}^{\tau\theta}\sigma''\left(z_{n}^{\theta}\right)\dot{z}_{n}^{\theta}\label{firstOvals}
\end{equation}
\begin{gather}
\frac{\partial\ddot{z}_{n+1}^{\tau}}{\partial z_{n}^{\theta}}=W_{(n+1)(n)}^{\tau\theta}\left(\sigma'''\left(z_{n}^{\theta}\right)\left(\dot{z}_{n}^{\theta}\right)^{2}+\right.\notag\\
\left.+\;\vphantom{\left(dot{z}_{n}^{\theta}\right)^{2}}\sigma''\left(z_{n}^{\theta}\right)\ddot{z}_{n}^{\theta}\right)\label{secondOvals}
\end{gather}
And similarly for derivatives of $E$ with respect to $\dot{z}_{n}^{\theta}$, however since $z_{n+1}^{\tau}$ does not depend on $\dot{z}_{n}^{\theta}$, only two products are left:
\begin{equation}
\frac{\partial E}{\partial\dot{z}_{n}^{\theta}}=\sum_{\tau}\frac{\partial E}{\partial\dot{z}_{n+1}^{\tau}}\frac{\partial\dot{z}_{n+1}^{\tau}}{\partial\dot{z}_{n}^{\theta}}+\frac{\partial E}{\partial\ddot{z}_{n+1}^{\tau}}\frac{\partial\ddot{z}_{n+1}^{\tau}}{\partial\dot{z}_{n}^{\theta}}\label{bpFirst}
\end{equation}
\begin{equation}
\frac{\partial\dot{z}_{n+1}^{\tau}}{\partial\dot{z}_{n}^{\theta}}=W_{(n+1)(n)}^{\tau\theta}\,\sigma'\left(z_{n}^{\theta}\right)\label{firstOfirst}
\end{equation}
\begin{equation}
\frac{\partial\ddot{z}_{n+1}^{\tau}}{\partial\dot{z}_{n}^{\theta}}=2\: W_{(n+1)(n)}^{\tau\theta}\,\sigma''\left(z_{n}^{\theta}\right)\dot{z}_{n}^{\theta}\label{secondOfirst}
\end{equation}
Finally since $\ddot{z}_{n}^{\theta}$ can only be encountered in expression for $\ddot{z}_{n+1}^{\tau}$:
\begin{equation}
\frac{\partial E}{\partial\ddot{z}_{n}^{\theta}}=\sum_{\tau}\frac{\partial E}{\partial\ddot{z}_{n+1}^{\tau}}\frac{\partial\ddot{z}_{n+1}^{\tau}}{\partial\ddot{z}_{n}^{\theta}}\label{bpSecond}
\end{equation}
\begin{equation}
\frac{\partial\ddot{z}_{n+1}^{\tau}}{\partial\ddot{z}_{n}^{\theta}}=W_{(n+1)(n)}^{\tau\theta}\sigma'\left(z_{n}^{\theta}\right)\label{secondOsecond}
\end{equation}
Derivatives of $E$ with respect to weights between $n$th and $n-1$st layers can be written as:
\begin{gather}
\frac{\partial E}{\partial W_{(n)(n-1)}^{\theta\kappa}}=\frac{\partial E}{\partial z_{n}^{\theta}}\sigma\left(z_{n-1}^{\kappa}\right)+\frac{\partial E}{\partial\dot{z}_{n}^{\theta}}\sigma'\left(z_{n-1}^{\kappa}\right)\dot{z}_{n-1}^{\kappa}+\notag\\
+\frac{\partial E}{\partial\ddot{z}_{n}^{\theta}}\left(\sigma''\left(z_{n-1}^{\kappa}\right)\left(\dot{z}_{n-1}^{\kappa}\right)^{2}+\sigma'\left(z_{n-1}^{\kappa}\right)\ddot{z}_{n-1}^{\kappa}\right)\label{previousWeightsFull}
\end{gather}
As for thresholds,
\begin{equation}\frac{\partial E}{\partial t_{n}^{\theta}}=\frac{\partial E}{\partial z_{n}^{\theta}}\label{thresh}\end{equation}
For initial connections between $a$ and $z_{1}^{\alpha}$
\[\frac{\partial E}{\partial W_{(1)(0)}^{\alpha\,1}}=\frac{\partial E}{\partial z_{1}^{\alpha}}a+\frac{\partial E}{\partial\dot{z}_{1}^{\alpha}}\]
Note that since $z_{1}^{\alpha}$ is a linear combination of $a$ and $b$ we can omit calculations of higher derivatives of $E$ with respect to $z_{1}^{\alpha}$ on the previous step. The whole procedure is as follows:
\begin{enumerate}
\item Forward propagation
	\begin{enumerate}
	\item Start with (\ref{initVals},\ref{initFirst},\ref{initSecond})
	\item Use (\ref{valsProp}) to obtain next layer values, (\ref{firstProp}) and (\ref{secondProp}) with their analogues for $b$ and (\ref{mixedProp}) to obtain derivatives. 
	\item Repeat (b) until the final layer is reached.
	\end{enumerate}
\item Backward propagation
	\begin{enumerate}
	\item Calculate cost function and its derivatives (\ref{eDerivLast}) with respect to output and derivatives of the output.
	\item Use (\ref{lastWeightsFull}) to calculate error derivatives with respect to last weights matrix.
	\item Use (\ref{bpVals}) with (\ref{valsOvals}), (\ref{firstOvals}) and (\ref{secondOvals}) to obtain the error derivatives with respect to previous layer values, (\ref{bpFirst}) with (\ref{firstOfirst}) and (\ref{secondOfirst}), (\ref{bpSecond}) with (\ref{secondOsecond}) for derivatives with respect to their partials.
	\item Use (\ref{previousWeightsFull}) to calculate error derivatives with respect to weights and (\ref{thresh}) to evaluate those with respect to thresholds.
	\item Repeat (c) and (d) until the input layer is reached.
	\end{enumerate}
\end{enumerate}
\section{N-dimensional case, higher orders}
\subsection{Forward propagation}
For expanding the algorithm further we will use vector notation for input variables $A=\left\{a,b,c,\dots\right\}$ and related derivatives as follows:
$\mathbb{D}^{S}$ is a differential operator, where $S$ is a vector denoting the order of derivative with respect
to each of the input variable. For example, $\mathbb{D}^{\left(1,0,3,\ldots\right)}=\nicefrac{\partial^{4}}{\partial a\partial c^{3}}$. Zero order operator is included, $\mathbb{D}^{\left(0,0,0,\ldots\right)}\equiv\hat{1}$.
In this notation for input layer:
\begin{equation}z_{1}^{\alpha}=\sum_{i}W_{(1)(0)}^{\alpha\: i}A_{i}\label{ndInitVal}\end{equation}
\begin{equation}\mathbb{D}^{I}z_{1}^{\alpha}=\frac{\partial z_{1}^{\alpha}}{\partial A_{i}}=W_{(1)(0)}^{\alpha\: i}\label{ndInit}\end{equation}
Where $I$ is a vector with 1 on position $i$ and 0 elsewhere. Any other first layer derivative is zero.
Error function will be written as: \[E\left(A,\mathbb{D}^{\left(0,0,0,\ldots\right)}z_{n+1}^{\tau},\mathbb{D}^{\left(1,0,0,\ldots\right)}z_{n+1}^{\tau},\ldots\right)\]
and forward pass formula turns into:
\begin{equation}\mathbb{D}^{s}z_{n}^{\theta}=\sum_{\kappa}W_{(n)(n-1)}^{\theta\kappa}\mathbb{D}^{s}\sigma\left(z_{n-1}^{\kappa}\right)\label{ndProp}\end{equation}
It is applied for every $s$ necessary to calculate the cost function. Note that higher derivatives can spawn a considerable amount of the lower order ones from previous layer. To estimate all terms required to calculate the error function one can use the following:

\begin{mylemma}
A expression for $n+1$st layer partial derivative ${\mathbb{D}}^{S}\sigma\left(z_{n}\right)$ can depend on two $n$th layer derivatives $\mathbb{D}^{P}z_{n}$ and $ \mathbb{D}^{Q}z_{n}$: \[{\mathbb{D}}^{S}\sigma\left(z_{n}\right)=f(\mathbb{D}^{P}z_{n},\mathbb{D}^{Q}z_{n},\ldots)\]
if, and only if its operator \:$\mathbb{D}^{S}$ can be presented as their product: 
\[\mathbb{D}^{S}=\mathbb{D}^{P}\mathbb{D}^{Q}\Leftrightarrow S=P+Q\]
where $P$ and $Q$ are vectors with non-negative integer components.
\end{mylemma}

\subsection{Backward propagation}
After feedforward procedure for set of all derivatives required to evaluate the cost function (0 order included) one should find its derivatives:
\begin{equation}\frac{\partial E}{\partial\mathbb{D}^{s}z_{n+1}}\label{ndErrorPartials}\end{equation}
Which are to be propagated backwards, and for every $\mathbb{D}^{r}$ operator:
\[\frac{\partial E}{\partial\mathbb{D}^{r}z_{n}^{\theta}}=\sum_{s}\frac{\partial{\mathbb{D}}^{s}\sigma\left(z_{n}^{\theta}\right)}{\partial\mathbb{D}^{r}z_{n}^{\theta}}\sum_{\tau}\frac{\partial E}{\partial\mathbb{D}^{s}z_{n+1}^{\tau}}W_{(n+1)(n)}^{\tau\theta}\]
Where $s$ runs through every operator that depends on $\mathbb{D}^{r}z_{n}^{\theta}$ as states lemma 1. Further simplifications will be made using the following:
\begin{mylemma}
If $\mathbb{D}^{S}$ can be decomposed to $\mathbb{D}^{Q}$$\mathbb{D}^{P}$ product then partial derivative of $\mathbb{D}^{S}\sigma\left(z_{n}\right)$ can be found as:
\begin{equation}\frac{\partial{\mathbb{D}}^{S}\sigma\left(z_{n}^{\theta}\right)}{\partial\mathbb{D}^{P}z_{n}^{\theta}}=C_{S}^{P}\frac{\partial}{\partial z_{n}}{\mathbb{D}}^{Q}\sigma\left(z_{n}\right)\label{ndLayerPartial}\end{equation}
where C is a combinatorial coefficient, equal to
\[C_{S}^{P}=\prod_{i}\frac{S_{i}!}{P_{i}!(S_{i}-P_{i})!}\]
\end{mylemma}

With lemma 2 we can rewrite backward pass formula with operators which were used for forward pass:
\begin{gather}
\frac{\partial E}{\partial\mathbb{D}^{r}z_{n}^{\theta}}=\sum_{s}\left(C_{s}^{r}\frac{\partial}{\partial z_{n}}\mathbb{D}^{s-r}\sigma(z_{n}) \cdot \vphantom{\sum_{\tau}} \notag\right.\\
\left. \cdot\sum_{\tau}\frac{\partial E}{\partial\mathbb{D}^{s}z_{n+1}^{\tau}}W_{(n+1)(n)}^{\tau\theta}\right)\label{ndBP}
\end{gather}
Where $s$ runs through every partial derivative used and for which $s-r$ got no negative components.
Derivatives of $E$ with respect to weights between $n$th and $n-1$st layers are calculated as follows:
\begin{equation}\frac{\partial E}{\partial W_{(n)(n-1)}^{\theta\kappa}}=\sum_{r}\frac{\partial E}{\partial\mathbb{D}^{r}z_{n}^{\theta}}{\mathbb{D}}^{r}\sigma\left(z_{n-1}^{\kappa}\right)\label{ndWeights}\end{equation}
Summation index r gathers every differential operator used.
Formula for thresholds remains the same.
\begin{equation}\frac{\partial E}{\partial t_{n}^{\theta}}=\frac{\partial E}{\partial z_{n}^{\theta}}\label{ndThresh}\end{equation}
Algorithm proceeds as follows: using initial conditions (\ref{ndInitVal}), (\ref{ndInit}) and formula (\ref{ndProp}) derivatives and values are propagated to the last layer, on which derivatives of $E$ (\ref{ndErrorPartials}) are calculated and propagated backwards using (\ref{ndBP}). Derivatives with respect to weights and thresholds are obtained from (\ref{ndWeights}) and (\ref{ndThresh}).

\section{Conclusion}
A universal algorithm for training neural networks with error function relaying on output derivatives is built. When applied to PDE, it creates a completely different technique with its unique advantages and disadvantages. Method can be used along with classical approaches. For example, it can improve the accuracy of a previously obtained solution. To do that one should first train a network to fit the previous solution and then to run presented procedure to shrink any remaining discrepancy. Implementation of batch mode training will require an additional index for all $z$ and their derivatives. Matrix-vector formulas will turn into matrix-matrix ones and expressions for weights updates previously written as outer products will become matrix products as well.


%

\appendices
\section{Fa\`a di Bruno's formula}
To prove lemmas, a combinatorial form\cite{21}: of the formula will be used:
\begin{equation}
\frac{\partial^{n}}{\partial x_{1}\cdots\partial x_{n}}f(z)=\!\sum_{\pi\in\Pi}f^{(|\pi|)}(z)\cdot\!\prod_{B\in\pi}\!\frac{\partial^{|B|}z}{\prod_{j\in B}\partial x_{j}}\label{faa}
\end{equation}
Where $\pi$ runs through the set of $\Pi$ of all partitions of $\left\{1\dots n\right\}$,
$B\in\pi$ states that B runs through the list of all of the blocks of the partition $\pi$.
$|\pi|$ denotes the number of blocks in partition $\pi$ and $|B|$ is the number of elements in block $B$.
\section{Proof of Lemma 1}
\begin{proof}
Suppose that in formula (\ref{faa}) we take a subset $\Pi_{2}$ of all partitions $\Pi$ with two blocks in every element, so that $\pi=\left\{ B_{1},B_{2}\right\}$. Union of each pair of those blocks is the set itself, and every possible choice of two blocks is presented. Vice versa, any block encountered in any element of $\Pi$ is a part of two-block partition $\Pi_{2}$ where the rest blocks are merged into one. $P$, $Q$ and $S$ will be the vectors that count appearances (or count-vectors, for brevity) of variables in $B_{1}$, $B_{2}$ and $\left\{1\dots n\right\}$, respectively. Note that one input variable can take more than one slot in $\left\{1\dots n\right\}$.
\end{proof}
\section{Proof of Lemma 2}
\begin{proof}
Consider set $\left\{1\dots n\right\}$ corresponding to $\mathbb{D}^{S}$, and set of all its partitions $\Pi_{S}$. Take all elements of $\Pi_{S}$ which contain a certain block $B$, then remove this block from each of selected element. Result will be a set of all partitions for $\left\{1\dots n\right\}\setminus B$. Same partition set will correspond to operator $\mathbb{D}^{S-P}$, provided that P is a count-vector for $B$. However, according to (\ref{faa}) the number of blocks will be greater by one than actual $|\pi|$ for partition of $\left\{1\dots n\right\}\setminus B$, since it contained the removed block. Thus, a partial derivative with respect to $z_{n}$ is required. Coefficient $C_{S}^{P}$ counts different choices of set $B$ which produce the same $P$ count-vector.
\end{proof}

\section{$\mathbb{D}^{S}$ operators}
We include a basic list of $\mathbb{D}^{S}$ operators. They can be symmetrically extended to three or more variables. For brevity the argument of $\sigma\left(z\right)$ is omitted.
\begin{gather*}
\mathbb{D}^{(2,2)}\sigma=\sigma''''z_{a}^{2}z_{b}^{2}+\sigma'''(4z_{ab}z_{a}z_{b}+z_{bb}z_{a}^{2}+z_{aa}z_{b}^{2})+\\
+\sigma''(2z_{ab}^{2}+2z_{a}z_{abb}+z_{aa}z_{bb}+2z_{aab}z_{b})+\sigma'z_{aabb}
\end{gather*}
\begin{gather*}
\mathbb{D}^{(3,1)}\sigma=\sigma''''z_{a}^{3}z_{b}+\sigma'''(3z_{aa}z_{a}z_{b}+3z_{ab}z_{a}^{2})+\\+\sigma''(3z_{aa}z_{ab}+3z_{aab}z_{a}+z_{aaa}z_{b})+\sigma'z_{aaab}
\end{gather*}
\begin{gather*}
\mathbb{D}^{(2,1)}z_{n}^{\theta}=\sigma'''z_{a}^{2}z_{b}+\sigma''(2z_{a}z_{ab}+z_{aa}z_{b})+\sigma'z_{aab}
\end{gather*}
\[\mathbb{D}^{(1,1)}\sigma=\sigma''z_{a}z_{b}+\sigma'z_{ab}\]
\begin{gather*}
\mathbb{D}^{(4,0)}\sigma=\sigma''''z_{a}^{4}+6\sigma'''z_{a}^{2}z_{aa}+\\+\sigma''(3z_{aa}^{2}+4z_{a}z_{aaa})+\sigma'z_{aaaa}
\end{gather*}
\[\mathbb{D}^{(3,0)}\sigma=\sigma'''z_{a}^{3}+3\sigma''z_{aa}z_{a}+\sigma'z_{aaa}\]
\[\mathbb{D}^{(2,0)}\sigma=\sigma''z_{a}^{2}+\sigma'z_{aa}\]

\section*{Acknowledgment}
Author expresses a great appreciation to E.A.~Dorofeev and Y.N.~Sviridenko for making this research possible. Sincere gratitude for helping with material if offered to A. Haas. A special thanks is extended to S.M.~Bosnyakov for providing a very valuable criticism.

\ifCLASSOPTIONcaptionsoff
  \newpage
\fi

\end{document}